\documentclass[letterpaper, 10pt, conference]{ieeeconf}
\IEEEoverridecommandlockouts
\usepackage{cite}
\usepackage{amsmath,amssymb,amsfonts}
\usepackage{algorithmic}
\usepackage{graphicx}
\usepackage{textcomp}
\usepackage{xcolor}
\usepackage{gensymb}
\usepackage{flushend}
\usepackage{booktabs}
\usepackage{xcolor}
\usepackage{icomma}
\usepackage{dblfloatfix} 
\usepackage{adjustbox}
\usepackage{tikz}
\usepackage[font=footnotesize]{subcaption}
\usepackage{todonotes}
\usepackage{makecell}
\usepackage{siunitx}

\usepackage{fancyhdr}

\fancypagestyle{IEEECopyright}{
	\lfoot{\copyright2025 IEEE. Accepted at 2025 IEEE International Conference on Robotics and Automation (ICRA).\\
  DOI: 10.1109/ICRA55743.2025.11127847%
  } 
}

\definecolor{mygreen}{RGB}{0,127,20} 
\def\BibTeX{{\rm B\kern-.05em{\sc i\kern-.025em b}\kern-.08em
		T\kern-.1667em\lower.7ex\hbox{E}\kern-.125emX}}

\title{\bf A Helping (Human) Hand in Kinematic Structure Estimation}

\begin{document}
	
\author{Adrian Pfisterer$^{*,1,2}$ \qquad Xing Li$^{*,1,2}$ \qquad Vito Mengers$^{1,2}$ \qquad Oliver Brock$^{1,2}$
\thanks{$^{*}$ Both authors contributed equally to this work.}%
\thanks{$^{1}$ Robotics and Biology Laboratory, Technische Universität Berlin}
\thanks{$^{2}$ Science of Intelligence, Research Cluster of Excellence, Berlin}
\thanks{We gratefully acknowledge funded provided by the Deutsche Forschungsgemeinschaft (DFG, German Research Foundation) under Germany’s Excellence Strategy – EXC 2002/1 “Science of Intelligence” – project number 390523135.}}
\maketitle

\begin{abstract}
Visual uncertainties such as occlusions, lack of texture, and noise present significant challenges in obtaining accurate kinematic models for safe robotic manipulation. We introduce a probabilistic real-time approach that leverages the human hand as a prior to mitigate these uncertainties. By tracking the constrained motion of the human hand during manipulation and explicitly modeling uncertainties in visual observations, our method reliably estimates an object’s kinematic model online. We validate our approach on a novel dataset featuring challenging objects that are occluded during manipulation and offer limited articulations for perception. The results demonstrate that by incorporating an appropriate prior and explicitly accounting for uncertainties, our method produces accurate estimates, outperforming two recent baselines by \SI[detect-weight]{195}{\percent} and \SI[detect-weight]{140}{\percent}, respectively. Furthermore, we demonstrate that our approach's estimates are precise enough to allow a robot to manipulate even small objects safely.
\end{abstract}


\thispagestyle{IEEECopyright}

\section{Introduction}

Safe and reliable robotic manipulation with articulated objects requires knowledge of their kinematic models. A common approach for obtaining kinematic models involves visually tracking the motion of individual object parts relative to one another. However, this method is susceptible to uncertainties in the visual input due to factors like a lack of texture and occlusions, which hinder accurate observation and tracking of object-part motion. Additionally, noise in measurements can lead to outliers that significantly deviate from the object's true motion, skewing model estimates.

\begin{figure}[t]
	\centering
	\includegraphics[width=\columnwidth]{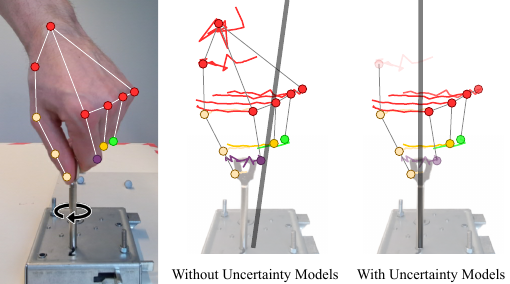} 
	\caption{Our approach estimates kinematic models from human hand motions for challenging objects where tracking object-part motions is difficult due to occlusions and a lack of texture (left). By deliberately modeling uncertainties based on hand properties, our approach weights observations and rejects outliers resulting from uncertainties and noisy measurements (center), significantly improving the estimation results (right).}
	\label{fig:title_figure}
\end{figure}

To address these challenges, we propose a probabilistic real-time approach that leverages the human hand as a perceptual prior to mitigate uncertainties in visual observations for accurate kinematic model estimation. The key idea is that when humans manipulate objects, their hand motions are partially constrained by the object's kinematic model. This allows us to infer object motion by tracking the hand, even when the object is fully occluded. In addition, the human hand's consistent structure---particularly its landmarks such as fingers and joints---facilitates reliable motion tracking. Moreover, we introduce novel uncertainty models based on hand properties, which quantify the uncertainty of individual landmarks and reject outliers unrelated to object motion, significantly improving the estimation, as shown in Fig.~\ref{fig:title_figure}.

To validate our method, we collected a novel benchmark dataset of ten objects. In particular, we include challenging objects like keys and slide locks, which have small articulations and are often heavily occluded during manipulation. To the best of our knowledge, visually estimating kinematic models of these small objects has not been explored yet. Our experiments demonstrate that the proposed method consistently produces accurate estimates for these challenging objects in an online manner, outperforming two recent baselines by \SI{195}{\percent} and \SI{140}{\percent}. These improvements stem from leveraging the human hand as a perceptual prior to enhance motion tracking and deliberately modeling uncertainties. We also demonstrate that our method's estimations are precise enough to allow a robot to reliably manipulate small objects, where minor estimation errors could cause significant forces that prevent successful manipulation.

\section{Related Work}
\label{sec:related work}

Our work leverages the human hand as a prior for robust Kinematic Structure Estimation (KSE). Therefore, this section reviews prior works on KSE, followed by approaches that learn manipulation policies from human hand motions. 

\subsection{Kinematic Structure Estimation}
\label{sec:kse}

Prior works on KSE often estimate the kinematic models of articulated objects based on the relative motions of object parts. Some approaches assume known trajectories of part poses~\cite{hausmanActiveArticulationModel2015a, JainLearnHybridObjKineFor2020, sturmAProbFrameLearnKin2011, LiuLearnArticConstFromOne2019}, while others track salient image features~\cite{martin-martinCoupledRecursiveEstimation2022, yewRPMNetRobustPoint2020, katz2008manipulating} or use learned motion maps clustered into pose trajectories~\cite{heppertCategoryIndependentArticulatedObject2022a}. Combining image features with shape-based segmentation can also assist in tracking object parts~\cite{mengersCombiningMotionAppearance2023, martinmartin2016AnIntegrated}. Additionally, some methods learn kinematic structures end-to-end from image sequences~\cite{jainScrewNetCategoryIndependentArticulation2021b}. Most approaches assume complete visibility of the object in order to estimate object-part motions. However, when humans manipulate objects---especially small ones as shown in Fig.~\ref{fig:title_figure}---they can often be largely occluded during the manipulation. Our approach circumvents this problem by tracking human hand motions, which are constrained by the object’s kinematics, to infer the kinematic model. Additionally, the distinct landmark features of human hands allow for reliable motion tracking. 

Many KSE methods provide deterministic estimates~\cite{hausmanActiveArticulationModel2015a, li2022learning, yewRPMNetRobustPoint2020, heppertCategoryIndependentArticulatedObject2022a, jainScrewNetCategoryIndependentArticulation2021b}, or are only partially probabilistic~\cite{jainDistributionalDepthBasedEstimation2022, arduengoRobustAdaptiveDoor2021}, making it difficult to account for uncertainty. This limitation is particularly problematic for small objects, where noise and occlusions can introduce significant uncertainty, compromising the estimation results. Additionally, few methods operate online~\cite{buchananOnlineEstimationArticulated2023, martin-martinCoupledRecursiveEstimation2022}, limiting their applicability in real-time tasks. Our work introduces a probabilistic real-time approach that effectively quantifies uncertainty based on hand properties, leading to enhanced estimation quality under noisy measurements. 

\subsection{Leanring Manipulation Policies from Hand Motions}
\label{sec:observing_human_manip}

Learning robot manipulation policies from human hand motions has recently gained significant interest~\cite{shaw2024LearningDexFrom}. Our work is related to approaches that directly track human hands to learn object affordances~\cite{damen2022rescaling, liu2022joint, liu2020forecasting, shan2020understanding, goyal2022human, mandikal2022dexvip}. However, these approaches do not provide kinematic models of the object, which is crucial for safe interaction, especially for small delicate objects. 

Our approach builds on existing methods that track human hand motions during manipulation to infer object kinematics for robotic manipulations. Existing approaches, however, track only a single point~\cite{regal2023usingSingle} or treat the hand as a rigid body~\cite{bahety2024screwmimic}, which often results in noisy estimations and require extensive fine-tuning on the robot to enhance estimation quality. In contrast, our approach demonstrates that by tracking individual hand landmarks and modeling their uncertainties using hand-specific properties, we can achieve accurate estimations that enable direct robot interaction without the need for additional interactive fine-tuning.

\section{Recursive Kinematic Model Estimation}

\begin{figure}[t]
	\centering
	\includegraphics[width=\columnwidth]{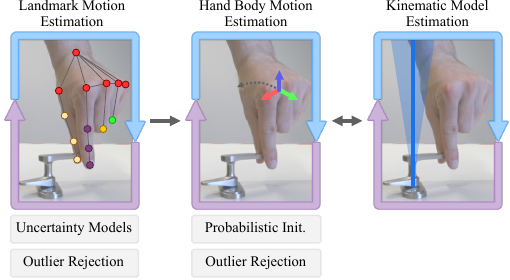}
	\caption{We decompose the estimation problem into three hierarchical levels, similar to the OMIP system~\cite{martin-martinCoupledRecursiveEstimation2022}. This hierarchical structure allows us to model uncertainties and reject outliers at two different levels, each grounded in physical priors related to hand properties.}
	\label{fig:method_overview}
    \vspace{-0.7em}
\end{figure}

Our method estimates kinematic models by tracking human hand motions. Inspired by the Online Multimodal Interactive Perception~(OMIP) system \cite{martin-martinCoupledRecursiveEstimation2022}, we decompose the estimation problem into three hierarchical levels: hand landmark motion, hand rigid body motion, and kinematic model estimation. At each level, we employ recursive Bayes filters. This hierarchical system structure allows us to model uncertainties and reject outliers at different levels of abstraction, as illustrated in Fig.~\ref{fig:method_overview}. 

While our approach builds upon the OMIP system, it addresses a crucial limitation: the assumption that objects are not occluded and offer sufficient image features for motion tracking. The following section will explain how to address this limitation by incorporating human hand motions into the estimation process.

\subsection{Landmark Motion Estimation}
\label{sec:tracking_landmarks}

\subsubsection{3D Hand Landmark Detection}

We need a method that detects and tracks hand motions in real-time while allowing for uncertainty modeling. Many existing approaches estimate full hand meshes from RGB images, but they often struggle with uncertainties in visual observations due to hand self-occlusions and are computationally expensive. Furthermore, modeling uncertainties in hand mesh estimations is challenging, particularly when these uncertainties arise from both observational data and the estimation models themselves~\cite{kendall2017uncertainties}. We address these challenges by estimating the motions of individual hand landmark features instead of full meshes. This increases computational efficiency and enables uncertainty modeling.

We detect hand landmarks by fusing two methods: AlphaPose~\cite{fang2023AlphaPose} and MediaPipe~\cite{lugaresi2019mediapipe}. Specifically, we first use MediaPipe to detect hand bounding boxes, which are then passed to AlphaPose to identify 21 hand landmarks along with their visibility scores $B^\text{vis}_i$. This fusion leverages the strengths of both methods: MediaPipe excels in detecting human hands~\cite{docekal2022human}, while AlphaPose provides confidence scores that allow us to track only visible hand landmarks. We back-project the detected landmarks into 3D using depth information and ignore them if their visibility scores are lower than a predefined threshold. 

\subsubsection{Estimating Landmark Motion}
\label{sec:method_feature_motion_kf}

We employ a Kalman Filter (KF) to estimate both the state and uncertainty of individual hand landmarks. The state of the KF is a vector $\mathbf{x}_t^{\text{lm}} = \begin{pmatrix} \mathbf{l}_t^{\text{lm}}& \mathbf{m}_t^\text{lm} \end{pmatrix}^T \in \mathbb{R}^{6}$ that contains the current estimate of a landmark's locations $\mathbf{l}_t^\text{lm} \in \mathbb{R}^{3}$ and its velocity $\mathbf{m}_t^\text{lm} \in \mathbb{R}^{3}$. The measurements used by the KF are 3D hand landmarks points $\mathbf{z}_t^\text{lm} \in \mathbb{R}^{3}$.

Since we track landmark motions over short time intervals, it is reasonable to assume a constant velocity, even though velocities may vary slightly during interactions. Therefore, our motion model predicts the next landmark state using $\Delta t$ along with the current location and velocity:
\begin{align}
	 \begin{pmatrix} \hat{\mathbf{l}}_{t+1} & \hat{\mathbf{m}}_{t+1} \end{pmatrix}^T = \begin{pmatrix} \mathbf{l}_{t} + \Delta t \, \mathbf{m}_{t} & \mathbf{m}_{t} \end{pmatrix}^T. 
\end{align}
As we can directly observe the landmark positions, our measurement model of the KF is:
\begin{align}
	 \mathbf{z}_t^\text{lm} = \begin{pmatrix} \mathbf{I}_{3} & \mathbf{0}_{3} \end{pmatrix} \hat{\mathbf{x}}_t^\text{lm} + \mathbf{v}_t,
\end{align}
where $\mathbf{v}_t$ is zero mean Gaussian noise.  

We now explain how we detect outliers resulting from noisy measurements and recognize landmarks lost or falsely predicted by the detection model. We compute the Mahalanobis distance \(M_t^\text{lm}\) between the predicted and measured observations:
\begin{equation}
	\label{eq_maha}
	M_t^\text{lm} = \sqrt{(\mathbf{z}_t^\text{lm} - \hat{\mathbf{z}}_t^\text{lm})^T (\mathbf{P}^{\hat{\mathbf{z}}}_t)^{-1} (\mathbf{z}_t^\text{lm} - \hat{\mathbf{z}}_t^\text{lm})}.
\end{equation}
Here, \(\mathbf{P}^{\hat{\mathbf{z}}}_t\) is the innovation covariance, and $\hat{\mathbf{z}}_t^\text{lm}$ denotes the predicted measurement derived from the prior (predicted) state estimate. If the filter receives a measurement and the corresponding \(M_t^\text{lm}\) exceeds a predefined threshold, we consider it an outlier and skip the correction step. Skipping the correction step leads to an increase in the landmark's covariance. If the covariance surpasses a predefined limit, the landmark is considered lost, and its filter is removed.

\subsubsection{Feature Saliency Uncertainty Model}

The saliency uncertainty model considers how well each landmark can be localized based on the quality of its image features. Although the visibility scores predicted by AlphaPose~\cite{fang2023AlphaPose} serve a similar purpose, they are poorly calibrated and cannot be used as probabilistic measurements for uncertainty modeling—a common issue with classifiers~\cite{minderer2021revisiting, guo2017calibration, muller2019does}. Instead, we propose a saliency-based model that assigns uncertainty scores for landmarks and integrates them into an uncertainty model based on a Kalman filter. 

Our saliency-based uncertainty model is based on the fact that vision models predict hand features using learned image features, and landmarks with lower visual saliency—those that are less distinct or prominent—are harder to localize accurately. For instance, the human wrist, with its large surface area, is difficult to localize consistently and precisely, whereas the tip of a finger is easier to identify due to its distinct appearance. Based on this observation, we design an uncertainty score $s_k \in \mathbb{R}$ for each type of hand landmark. We exclude the wrist joint entirely due to excessive uncertainty in its estimates. For all thumb joints, we assign a score of $s_\text{thmb} = 1.5$. Starting from the knuckle joints (MCP), we set $s_\text{mcp} = 1.5$. For the first finger joint (PIP) and second finger joint (DIP), we assign $s_\text{pip}=1.0$ and $s_\text{dip}=1.5$, respectively. Lastly, we set $s_\text{tip} = 0.5$ for all fingertips. These scores act as linear factors that scale the uncertainty estimated by the next uncertainty model that we will introduce in the following. 

\subsubsection{Motion Continuity Uncertainty Model}

We use a Kalman filter as both a state estimator and uncertainty model, grounded in the prior of motion continuity. We assume Gaussian-distributed uncertainty in continuous hand motion. By alternating between prediction and update phases, the KF refines state estimates while quantifying its uncertainty. 

We combine this motion continuity uncertainty model with the previously explained saliency uncertainty model to compute the final uncertainties for landmarks. Specifically, the covariance estimated by the KF, $\mathbf{P}^\text{lm}_t$, is scaled by scalar saliency factors $s_k$ to produce the adjusted covariance $\mathbf{R}^\text{rb}_t$:
\begin{align}
	\mathbf{R}^\text{rb}_t = s_k \mathbf{P}^\text{lm}_t,
\end{align}
which is then used for the next level---probabilistic hand body initialization and motion estimation.

\newcommand{\imgwidth}{0.17\textwidth}
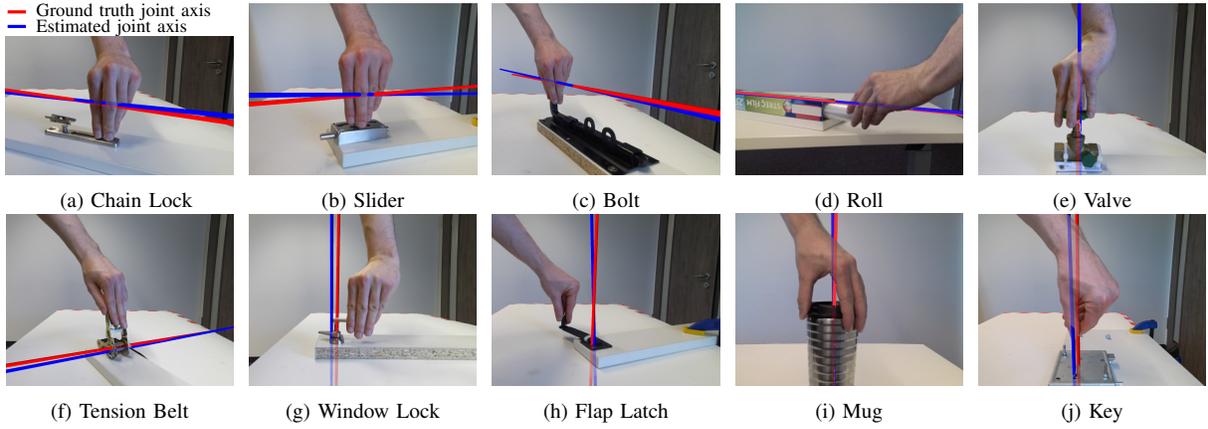
\begin{figure*}[ht]
	\centering
    \hspace{-0.9em}
	\subfloat[Chain Lock]{
		\begin{tikzpicture}
			\node[anchor=south west,inner sep=0] (image) at (0,0) {
				\adjincludegraphics[width=\imgwidth, trim={{.3\width} {.08\height} {.1\width} {0.12\height}},clip]{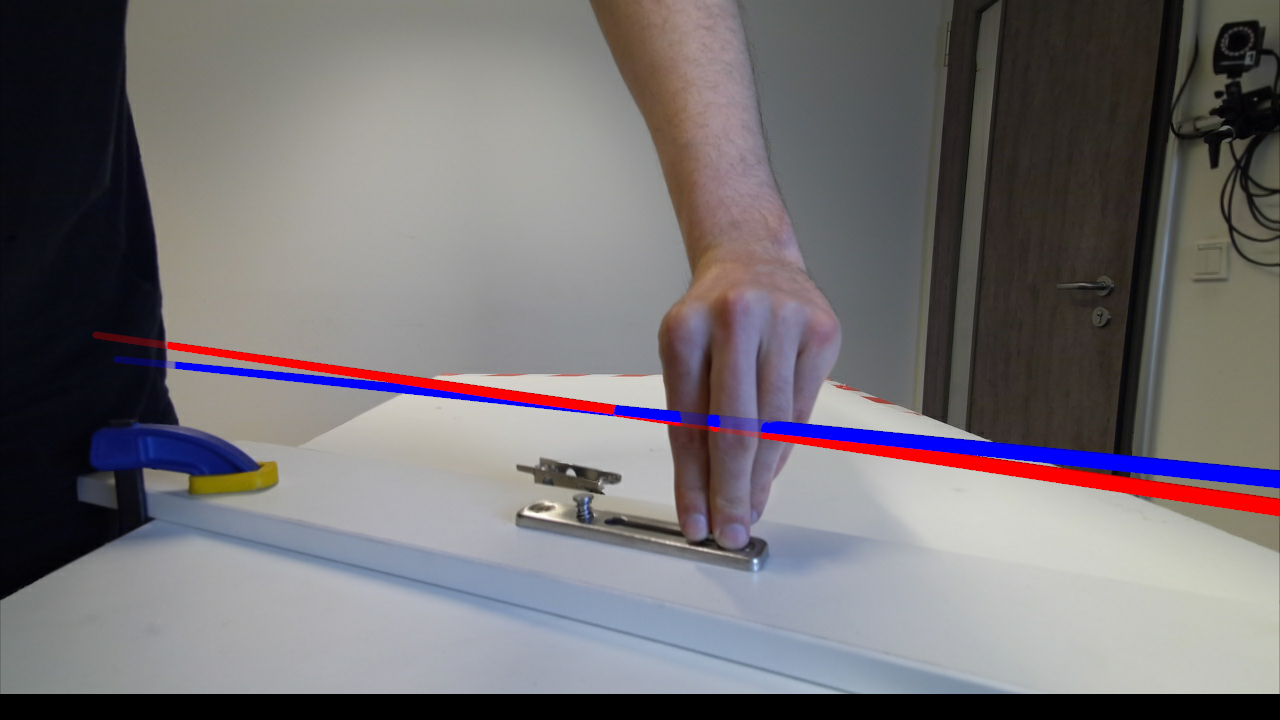}
			};
			\begin{scope}[x={(image.south east)},y={(image.north west)}]
				\fill [white] (0.0,0.81) rectangle (1.1,1.0);
				\node[anchor=west, font=\scriptsize] at (0.095,0.95) {Ground truth joint axis};
				\draw[red, line width=1.2pt] (0.01, 0.95) -- (0.09, 0.95); 
				\node[anchor=west, font=\scriptsize] at (0.095,0.86) {Estimated joint axis};
				\draw[blue, line width=1.2pt] (0.01, 0.865) -- (0.09, 0.865); 
			\end{scope}
		\end{tikzpicture}
	}
	\hspace{-0.8em}
	\subfloat[Slider]{\adjincludegraphics[width=\imgwidth, trim={{.25\width} {.08\height} {.15\width} {0.12\height}},clip]{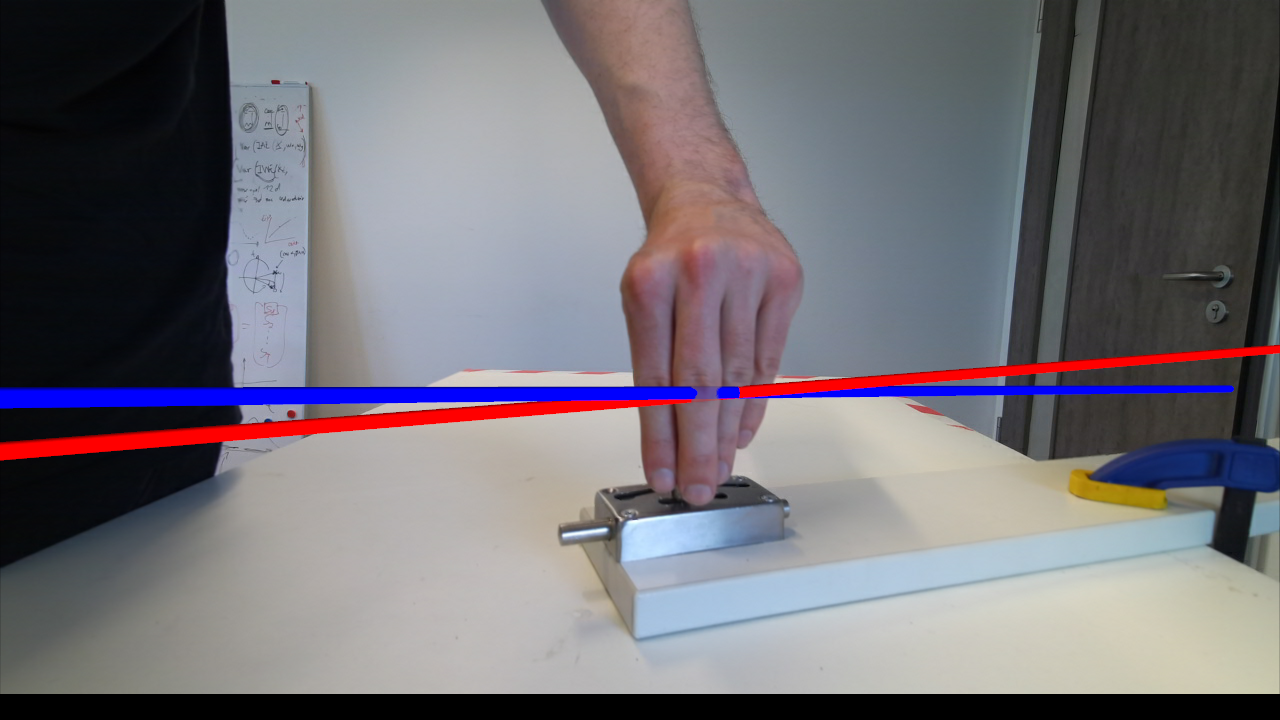}} \hspace{0.01\columnwidth}
	\subfloat[Bolt]{\adjincludegraphics[width=\imgwidth, trim={{.35\width} {.08\height} {.05\width} {0.12\height}},clip]{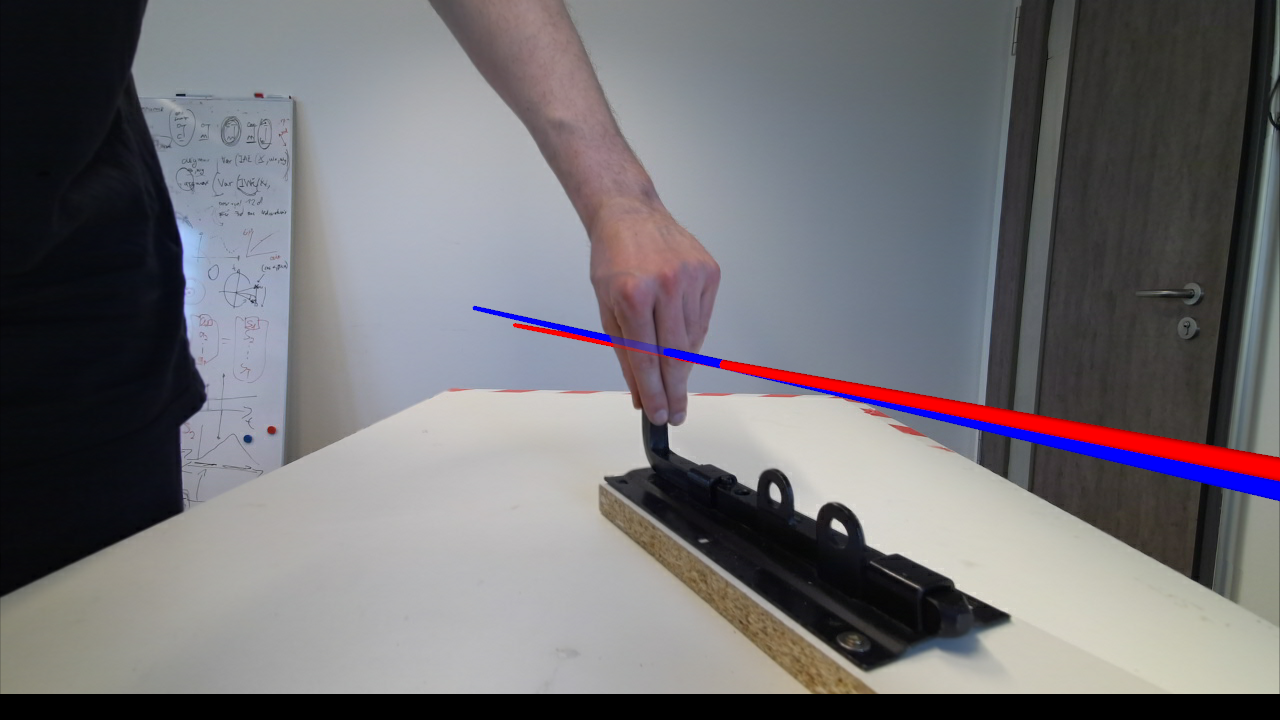}}\hspace{0.01\columnwidth}
	\subfloat[Roll]{\adjincludegraphics[width=\imgwidth, trim={{.1\width} {.15\height} {.3\width} {0.05\height}},clip]{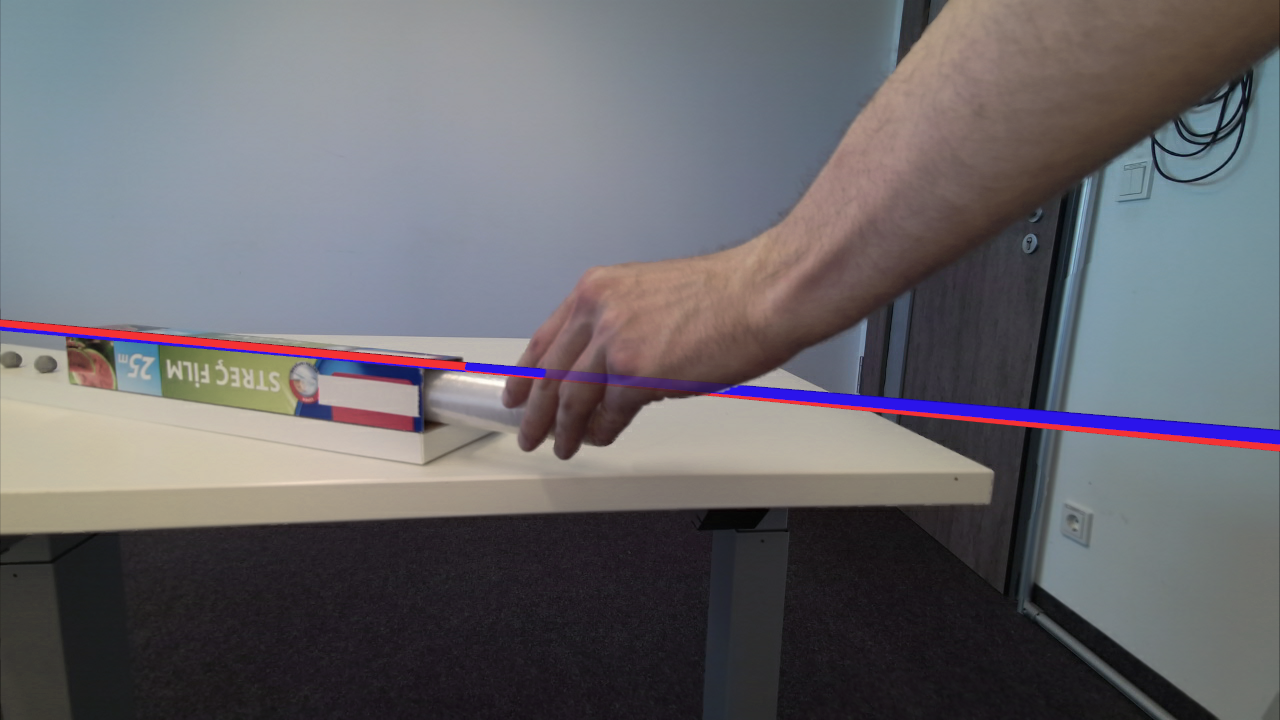}}\hspace{0.01\columnwidth}
	\subfloat[Valve]{\adjincludegraphics[width=\imgwidth, trim={{.3\width} {.12\height} {.1\width} {0.08\height}},clip]{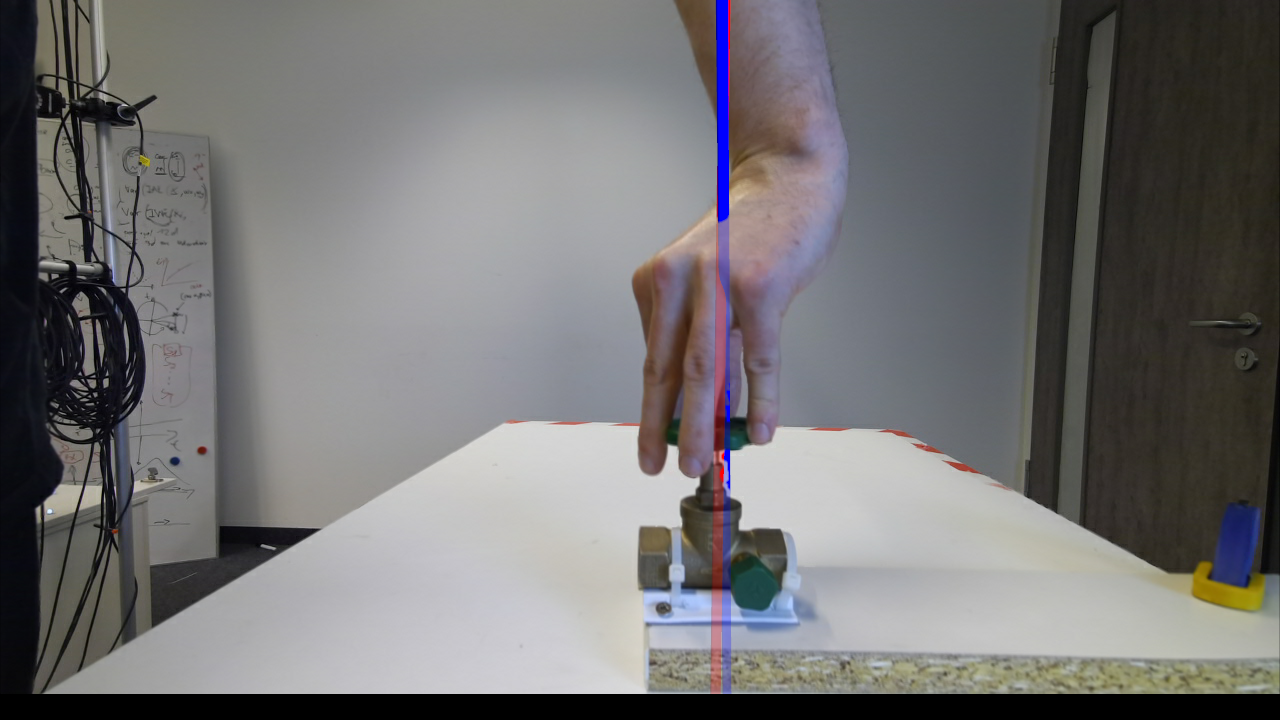}}\hspace{0.01\columnwidth}\
	\hspace{0.01\columnwidth}
	\subfloat[Tension Belt]{\adjincludegraphics[width=\imgwidth, trim={{.25\width} {.08\height} {.15\width} {0.12\height}},clip]{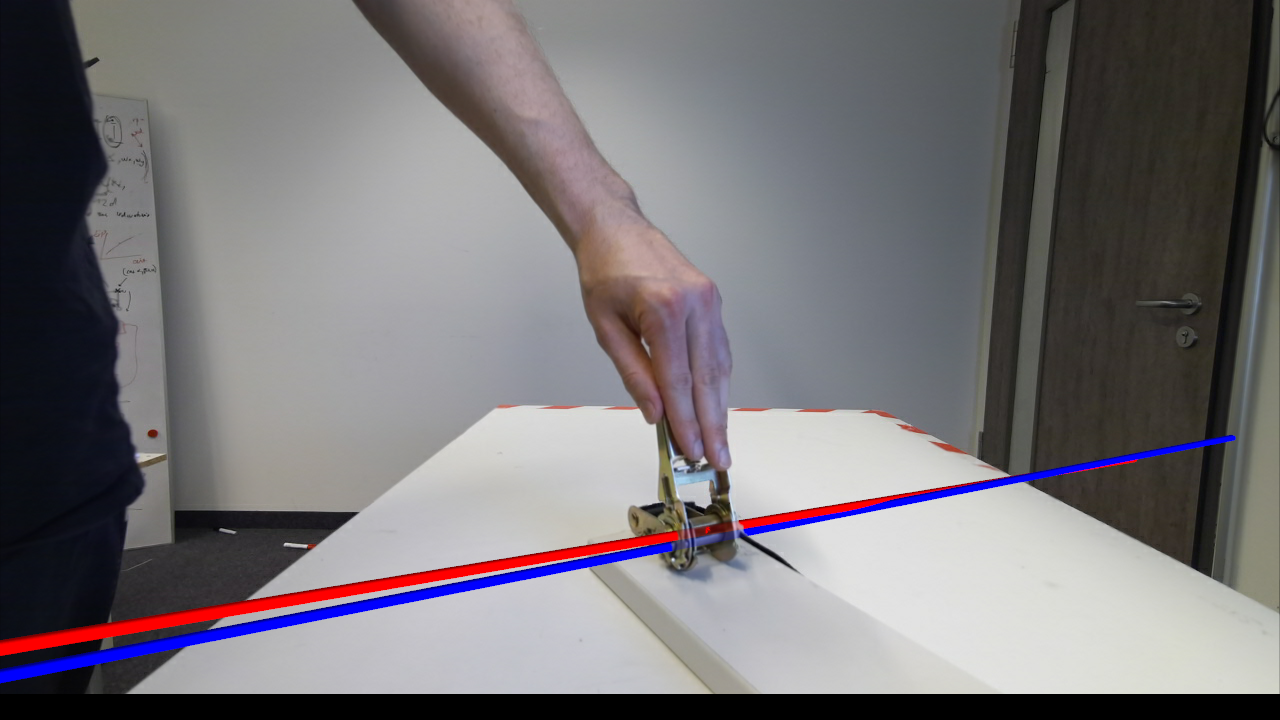}}\hspace{0.01\columnwidth}
	\subfloat[Window Lock]{\adjincludegraphics[width=\imgwidth, trim={{.26\width} {.08\height} {.14\width} {0.12\height}},clip]{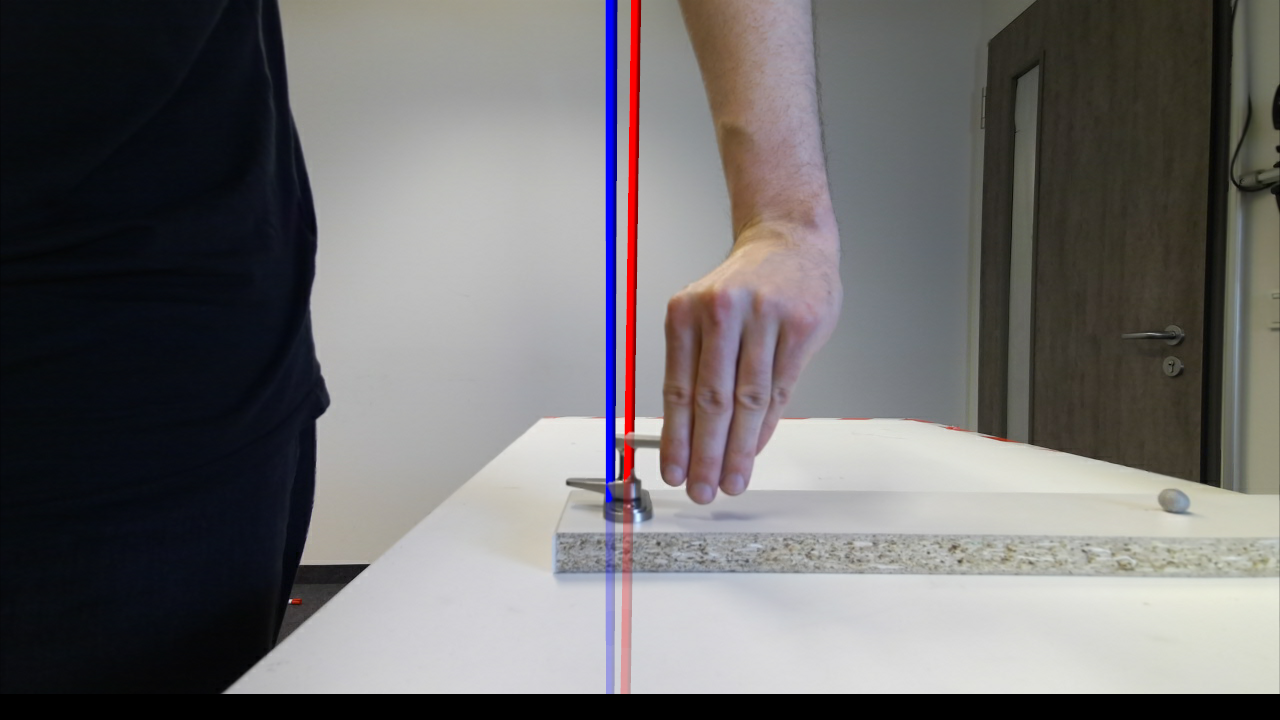}} \hspace{0.01\columnwidth}
	\subfloat[Flap Latch]{\adjincludegraphics[width=\imgwidth, trim={{.3\width} {.08\height} {.1\width} {0.12\height}},clip]{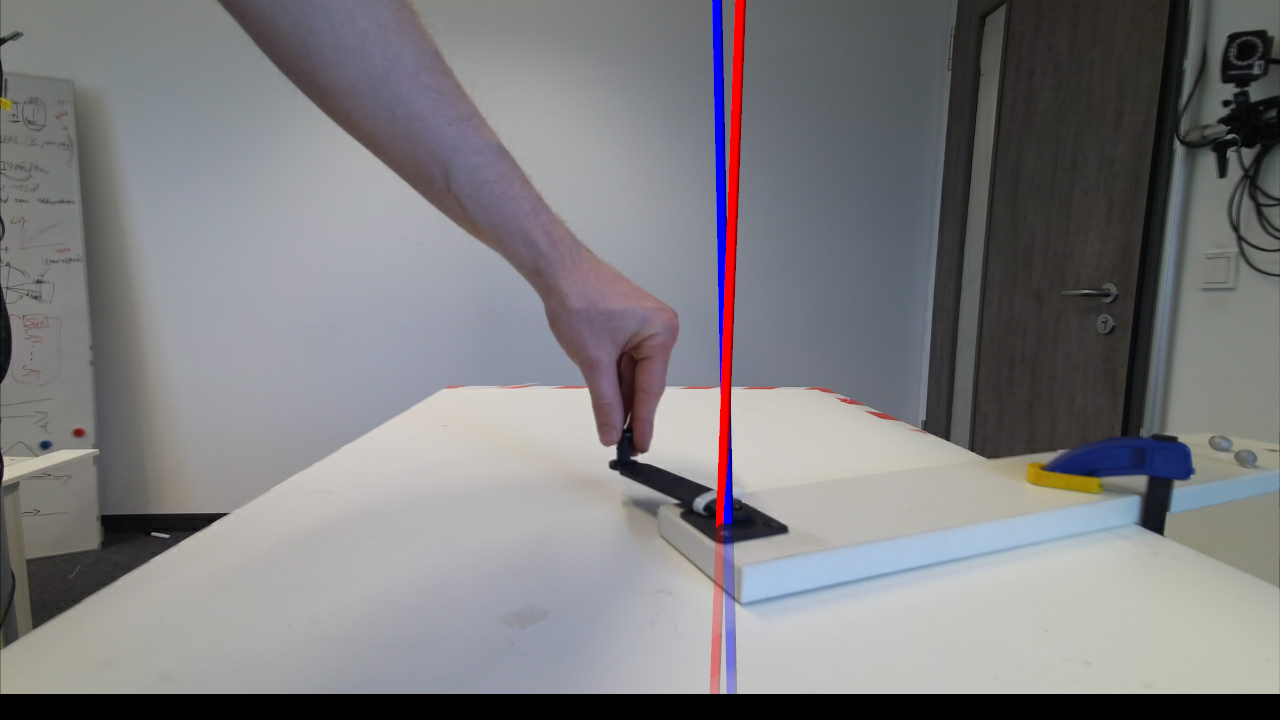}}\hspace{0.01\columnwidth}
	\subfloat[Mug]{\adjincludegraphics[width=\imgwidth, trim={{.25\width} {.08\height} {.25\width} {0.25\height}},clip]{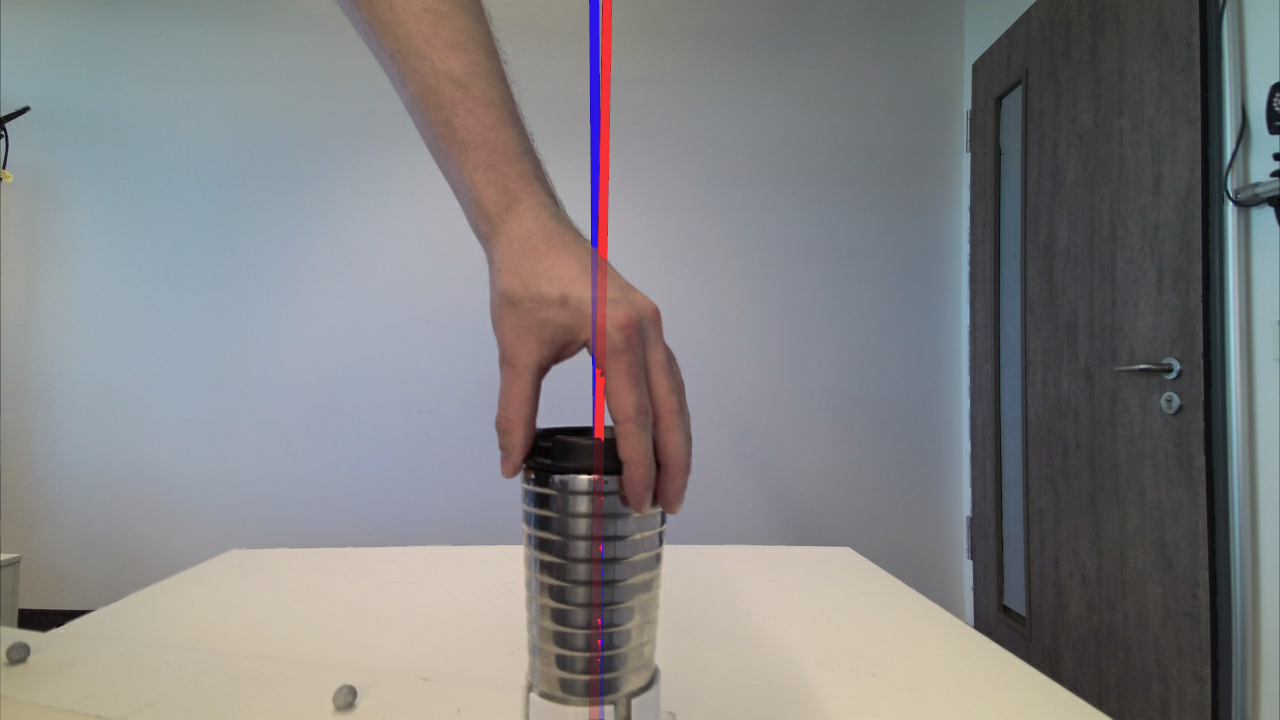}}\hspace{0.01\columnwidth} 
	\subfloat[Key]{\adjincludegraphics[width=\imgwidth, trim={{.2\width} {.1\height} {.2\width} {0.1\height}},clip]{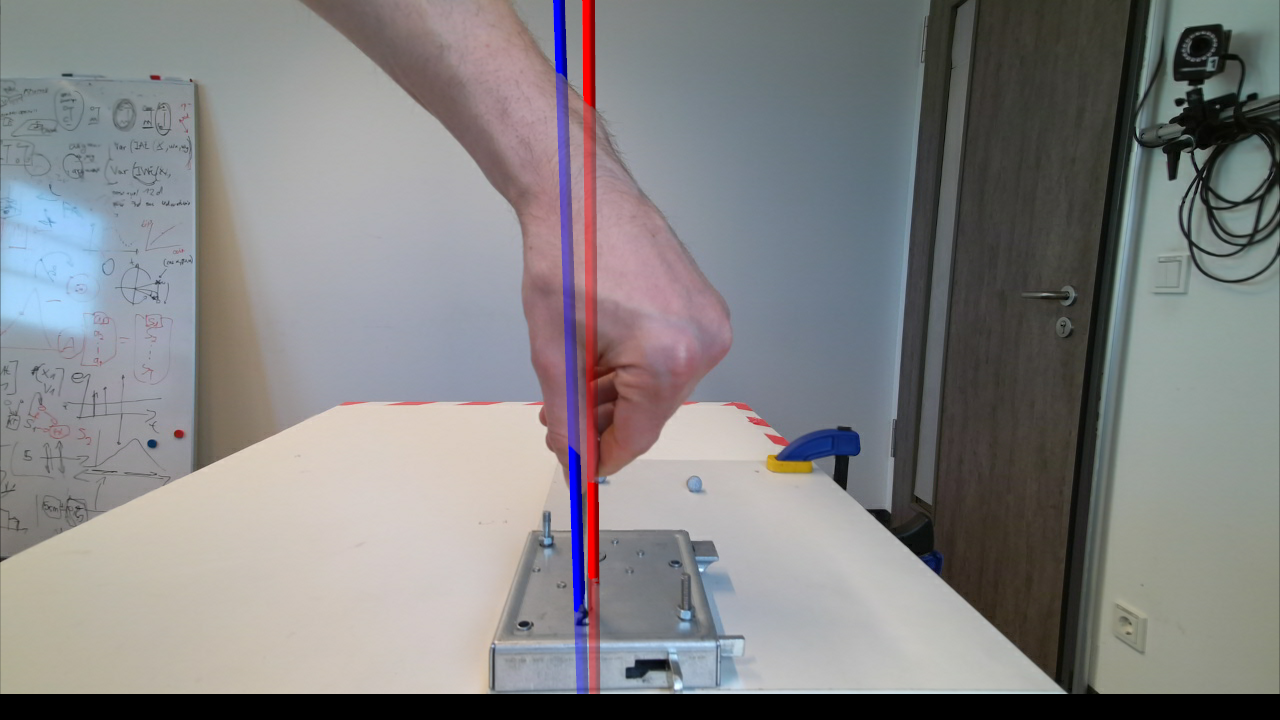}}
	\caption{Our method accurately estimates kinematic models of various challenging articulated objects from human hand motion in real-time.}
	\label{fig_artic_obj_dataset}

\end{figure*}

\subsection{Hand Body Motion Estimation}
\label{sec:pose_est}

By tracking individual hand landmark motions, we can estimate how the hand as a whole moves. The hand's motion is partially constrained by the kinematics of the manipulated object, and thus can be used to infer the object's kinematic structure. We treat the hand body as a 6D pose, i.e., rigid body, and estimate its motion accordingly. To achieve this, we first need to initialize a hand body pose based on the tracked trajectories of landmarks. 


Proper initialization of the hand pose is critical for accurate hand body motion estimation. If the landmarks for initialization deviate significantly from their subsequent motions, they can result in large innovations when updating the hand rigid body's state, causing drift in motion estimation. We introduce a \emph{probabilistic initialization} step that leverages the landmark uncertainties computed by our models. To find an initial hand body pose that best explains the observed motion, we use RANSAC~\cite{fenzi3DObjectRecognition2012}. We modify the algorithm to perform weighted sampling based on landmarks’ uncertainties, i.e., $\text{trace}(\mathbf{R}^\text{rb}_t)^{-1}$ to find an initial hand body transformation that best explains landmarks’ motions. Additionally, we weigh the residuals in the RANSAC algorithm and filter outliers, leading to robust hand body pose initialization. 

After initialization, we use an Extended Kalman Filter (EKF) to estimate the hand pose over time. The state of the EKF consists of the hand body pose $\mathbf{p}_t^\text{rb} \in \mathbb{R}^{6}$ and its velocity $\boldsymbol{\eta}_t^\text{rb} \in \mathbb{R}^{6}$ in exponential coordinates: $\mathbf{x}_t^\text{rb} = \begin{pmatrix} \mathbf{p}_t^\text{rb} & \boldsymbol{\eta}_t^\text{rb} \end{pmatrix}^T$. 

We utilize three motion models adopted from the OMIP~\cite{martin-martinCoupledRecursiveEstimation2022} system to perform the prediction step. The first model assumes that the pose remains unchanged. The second one predicts the next pose using the current pose, velocity, and the time elapsed. The last model uses the kinematic model estimated by the filter of the next level to predict a new pose and velocity. When measurements arrive, we compare all three predictions with the actual measurement and choose the most likely model for the correction step. By combining these three motion models, we can account for discontinuous human hand movements and leverage the current kinematic model belief to enhance hand body motion estimation. Furthermore, we compute the Mahalanobis distance $M^\text{rb}_t$ between the predicted landmark’s location and the measured one, similar to Equation~\ref{eq_maha}. Therefore, we can incorporate the physical prior about body motion continuity to refine the landmark motion estimation by rejecting outliers.

In summary, our approach estimates hand body motions from landmark motions while explicitly accounting for uncertainties and outliers at both levels. As we will demonstrate in the experiments (Section~\ref{sec:comp_baselines}), the resulting hand body motions allow us to accurately estimate kinematic models for challenging objects, whereas directly estimating hand poses from RGB images would be prone to errors due to visual uncertainties and computationally expensive.

\subsection{Kinematic Model Estimation}
\label{sec:kin_model_est}

We estimate the kinematic model of the object based on the hand body motions. We use the same EKFs as in the OMIP system~\cite{martin-martinCoupledRecursiveEstimation2022} and recursively estimate model type and parameters from the estimated hand body motions. For further details on the kinematic model estimation, we would like to refer the reader to the paper~\cite{martin-martinCoupledRecursiveEstimation2022}.

\section{Experiments}

\begin{table}
    \centering
    \begin{tabular}{l l l}
        \toprule
        Parameter & Value\\ 
        \midrule
        (E)KF initial cov. & \makecell[l]{$\mathbf{P}_0^\text{lm} = \text{blkdiag}(0.09\,\mathbf{I}_3, 0.06\,\mathbf{I}_3)$ \\ 
        $\mathbf{P}^\text{rb}_0 = \text{blkdiag}(0.05\,\mathbf{I}_3, 0.2\,\mathbf{I}_3, 0.1\,\mathbf{I}_3,  0.2\,\mathbf{I}_3)$\\ 
        $\mathbf{P}_0^\text{pris} = 3.0\,\mathbf{I}_4$  \\  $\mathbf{P}^\text{rev}_0 = \mathbf{I}_7$  } \\
        (E)KF process cov. & \makecell[l]{$\mathbf{Q}^\text{lm} = \text{blkdiag}(0.13\,\mathbf{I}_3  ,0.05\,\mathbf{I}_3 )^\dag$ \\ 
        $\mathbf{Q}^\text{rb} = \text{blkdiag}(0.75\,\mathbf{I}_3, 3\,\mathbf{I}_3, 2.4\,\mathbf{I}_3, 4.8\,\mathbf{I}_3)$ \\
        $\mathbf{Q}^\text{pris} = \text{blkdiag}(2.55\,\mathbf{I}_2, 0.7, 75)$ \\ 
        $\mathbf{Q}^\text{rev} = \text{blkdiag}(2.55\,\mathbf{I}_2, 0.3\,\mathbf{I}_3, 5.1, 75.0)$  } \\
        (E)KF meas. cov. & $\mathbf{R}^\text{lm} = 0.05\,\mathbf{I}_3^\dag$ \\
        Landmark unc. thresh. & $\text{trace}(\mathbf{P}^\text{lm}) \geq 0.3 ^\dag$\\
        Landmark vis. thresh. & $B^\text{vis}_i \geq 0.006^\dag$ \\
        Mahalanobis dist. thresh. \hspace{-2em} & $M^\text{lm}\geq0.19^\dag, \: M^\text{rb} \geq 0.25 ^\dag$\\
        \bottomrule
    \end{tabular}
    \caption{Hand-tuned and Bayesian-optimized~\cite{dewancker2016bayesian} parameters used in all experiments (\dag\ indicates Bayesian-optimized values)}
    \label{table:filter_param}
    \vspace{-0.8em}
\end{table}

We assess the performance of our approach for kinematic structure estimation across a variety of articulated objects. First, we describe the experimental setup and the evaluation metrics. Next, we demonstrate that our method achieves more accurate estimation results compared to two recent baselines, which do not account for uncertainties. We also show that incorporating uncertainty models significantly improves estimation performance. Lastly, we show that the estimated kinematic models are sufficiently precise to enable a robot to manipulate objects directly.

\subsection{Experimental Setup and Metrics}
\label{sec:exp_setup_met}

\subsubsection{Benchmark Dataset}

We collect a benchmark dataset for evaluation. We do so because we want to evaluate challenging scenarios, i.e., estimate kinematic models of small objects, which have not been thoroughly explored. Thus, we were unable to find a dataset that includes human manipulation of small articulated objects while providing ground truth kinematic models.

Fig.~\ref{fig_artic_obj_dataset} shows the ten articulated objects included in our dataset (four prismatic and six revolute objects). Three participants are tasked with manipulating these objects while maintaining constant contact. We record the manipulation sequences using a Microsoft Kinect Azure RGB-D camera and obtain ground truth data using a motion capture system.

\subsubsection{Evaluation Metric}
We utilize the tangent error metric $E_\text{tan}$ for evaluating the quality of estimates, as proposed by \mbox{Heppert et al.~\cite{heppertCategoryIndependentArticulatedObject2022a}}:
\begin{align}
\label{eq:tang_error}
	E_\text{tan} =  \frac{1}{q_\text{max}} \int_{0}^{q_\text{max}} \text{cos}^{-1} \left( \frac{\mathbf{t}_\text{gt}}{\lVert \mathbf{t}_\text{gt} \rVert } \cdot \frac{\mathbf{t}_\text{est}}{\lVert \mathbf{t}_\text{est} \rVert } \right) dq.
\end{align}
This metric quantifies the average deviation between the estimated tangent vector, $\mathbf{t}_\text{est}$, and the ground truth tangent vector, $\mathbf{t}_\text{gt}$, over an object's articulation range $q \in [0,\; q_\text{max}]$. The tangent vectors represent the direction of the admissible motion constrained by the kinematic models. We choose this metric as it provides a comprehensive measure of the estimated joint axis and indicates how well the robot can manipulate the object. For revolute joints, $q_\text{max}$ comes from the demonstration. For prismatic joints, the tangent remains constant, so the integral in Eq.~\ref{eq:tang_error} is omitted.

\subsubsection{Hyperparameter Tuning}

We divide our dataset into two subsets: one for hyperparameter tuning and the other for evaluation. We employ Bayesian hyperparameter optimization~\cite{dewancker2016bayesian} to fine-tune model parameters \emph{only} using four objects (valve, tension belt, window lock, and flap latch) manipulated by two of the three participants. For evaluation, we use unseen data consisting of manipulation sequences with all objects from a third participant. All parameters can be found in TABLE~\ref{table:filter_param}.

\subsection{Probabilistic Human Hand Tracking Enables Accurate Kinematic Model Estimation}
\label{sec:comp_baselines}

We first assess the effectiveness of our approach by comparing the accuracy of the estimated articulation models with that of two baseline methods. Both methods utilize human hand tracking to infer kinematic models but work offline. Additionally, we test a variant of our approach with both the uncertainty models and outlier rejection mechanisms disabled, to examine their impact on accuracy.

The first baseline method is a part of a larger system proposed by Regal et al.~\cite{regal2023usingSingle}. This method tracks a landmark on the hand and fits a kinematic model as a screw axis to its trajectory. While their system utilizes a mixed-reality headset for tracking, we obtain the landmark trajectory using our landmark motion estimation component. To ensure a fair comparison, we also apply outlier rejection to filter all landmark trajectories and select the one with most observations. The second baseline, ScrewMimic by Bahety et al.~\cite{bahety2024screwmimic}, fits a kinematic model to the motions of the right hand relative to the left hand. The hand motions are obtained using a hand pose estimator (FrankMocap~\cite{rong2021frankmocap}). Notably, Bahety et al. \cite{bahety2024screwmimic} introduce an iterative process that refines the estimated kinematic model using the robot's proprioception and force/torque measurements. However, since the other approaches do not involve a fine-tuning procedure, we only evaluate ScrewMimic's initial estimates. Additionally, we record a single hand motion relative to its initial pose for estimation. The authors of both baselines kindly provided their code, allowing for direct comparison. 

\begin{figure}
\vspace{-1em}
	\centering
	\includegraphics[width=\columnwidth]{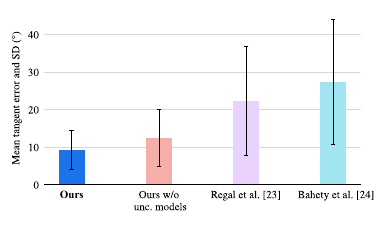}
	\caption{Compared to the two baseline methods, our approach achieves significantly higher accuracy and less variance on the evaluation dataset. Additionally, the results show that our uncertainty models further improve the accuracy of the estimations.}
	\label{fig:baseline_comp_fig}
	\vspace{-1em}
\end{figure}

Fig.~\ref{fig:baseline_comp_fig} shows the average tangent error over all 10 objects across the compared approaches. Our method substantially outperforms the others, with Regal et al.\cite{regal2023usingSingle} showing a \SI{140}{\percent} decrease in accuracy and Bahety et al.\cite{bahety2024screwmimic} a \SI{195}{\percent} decrease compared to our approach. Additionally, our method exhibits low variance, highlighting its robustness and consistency in estimating kinematic models. Without the uncertainty models, the accuracy decreases by \SI{34}{\percent}, and the variance increases, emphasizing the importance of modeling uncertainties for accurate estimates.

\begin{figure}
	\centering
	\begin{subfigure}[t]{0.48\columnwidth}
		\includegraphics[width=1\linewidth]{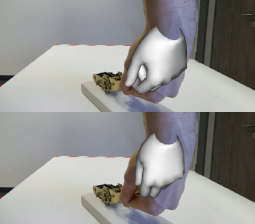}
		\caption{Bahety et al.~\cite{bahety2024screwmimic} using FrankMocap \cite{rong2021frankmocap}}
		\label{fig:result/failure_ours} 
	\end{subfigure}
	~
	\begin{subfigure}[t]{0.48\columnwidth}
		\includegraphics[width=1\linewidth]{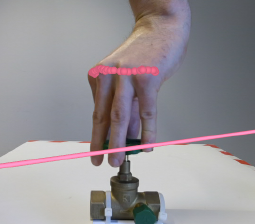}
		\caption{Regal et al.~\cite{regal2023usingSingle} tracking a single point}
		\label{fig:results/failure_screw_mimic}
	\end{subfigure}
	\caption{The baseline methods struggle to generate accurate estimations due to uncertainties in visual observations. (a): When manipulating the tension belt, partial occlusion of the hand makes it challenging to detect the hand pose directly from images, which affects the estimation accuracy of Bahety et al.~\cite{bahety2024screwmimic}. (b): Similarly, occlusions and the limited range of observed articulations permitted by the valve result in a short, noisy landmark trajectory. Combined with the lack of orientation data this leads to a poorly estimated rotation axis in the method of Regal et al.~\cite{regal2023usingSingle}.}
\label{fig:results/failure_case}
\vspace{-0.8em}
\end{figure}

Across all the objects evaluated, our method provides consistent accuracy, as shown in TABLE~\ref{tab:baseline_comp_tab}. In contrast, the baseline methods exhibit significant variation in performance across different objects. The discrepancy in the estimation quality lies in the uncertainty modeling. The baseline approaches rely on direct hand pose estimation or track only a single point to fit kinematic models. As a result, they are susceptible to visual uncertainties caused by noisy landmark observations and inaccurate hand pose estimates due to occlusions, as shown in Fig.~\ref{fig:results/failure_case}. Our approach, on the other hand, deliberately quantifies the uncertainty of individual landmarks based on hand properties, enhancing the robustness of the hand pose estimation against noisy measurements, leading to more accurate and consistent estimates, even for challenging objects with small articulations. 

\begin{table}
	\centering
	\begin{tabular}{l c c c c}
		\toprule
		\multicolumn{1}{c}{} & \multicolumn{4}{c}{Tangent error} \\
		\cmidrule{2-5}
		Object & \makecell{Regal \\ et al.\cite{regal2023usingSingle}} &  \makecell{Bahety \\et al.\cite{bahety2024screwmimic}} & \makecell{Ours w/o \\unc. models}  & \textbf{Ours} \\
		\midrule
		\multicolumn{4}{l}{\textit{Prismatic}} & \\	
		\quad Chain Lock & \SI{50}{\degree} & \SI{23}{\degree} & ~\textbf{\SI[detect-weight]{3}{\degree}} & ~\textbf{\SI[detect-weight]{3}{\degree}} \\
		\quad Slider & \SI{17}{\degree} & \SI{31}{\degree} & \SI{16}{\degree} & ~\textbf{\SI[detect-weight]{9}{\degree}} \\
		\quad Bolt & \SI{17}{\degree} & ~\textbf{\SI[detect-weight]{2}{\degree}} & \SI{12}{\degree} & ~\SI{7}{\degree} \\
		\quad Roll & ~\SI{3}{\degree} & ~\SI{3}{\degree} & ~\textbf{\SI[detect-weight]{1}{\degree}} & ~\textbf{\SI[detect-weight]{1}{\degree}} \\
		\midrule
		\multicolumn{4}{l}{\textit{Revolute}} & \\		
		\quad Valve & \SI{20}{\degree} & \SI{55}{\degree} & \SI{18}{\degree} & \textbf{\SI[detect-weight]{14}{\degree}} \\
		\quad Tension Belt & \SI{11}{\degree} & \SI{27}{\degree} &  ~\textbf{\SI[detect-weight]{7}{\degree}} & ~~\SI{8}{\degree} \\
		\quad Window Lock & \SI{49}{\degree} & \SI{23}{\degree} & \SI{25}{\degree} & \textbf{\SI[detect-weight]{16}{\degree}} \\
		\quad Flap Latch & \SI{15}{\degree} & \SI{27}{\degree} &  ~\textbf{\SI[detect-weight]{7}{\degree}}  & ~~\textbf{\SI[detect-weight]{7}{\degree}} \\
		\quad Mug & \textbf{\SI[detect-weight]{16}{\degree}} & \SI{28}{\degree} & \SI{22}{\degree} & \textbf{\SI[detect-weight]{16}{\degree}} \\
		\quad Key & \SI{26}{\degree} & \SI{55}{\degree} & \SI{14}{\degree} & \textbf{\SI[detect-weight]{13}{\degree}} \\
		\bottomrule
	\end{tabular}
	\caption{Both baseline methods exhibit considerable variance in estimation accuracy across various objects, while our approach consistently delivers small tangent errors and maintains robust performance. Tangent errors are rounded to the nearest degree, as sub-degree discrepancies are likely attributable to noise in the motion capture system used for recording ground truth data.}
	\label{tab:baseline_comp_tab}
\end{table}

\subsection{Direct Transfer to Real-World Robot Manipulation}
The previous experiments have demonstrated that our approach generates accurate kinematic models. We now show that the models estimated using our approach are precise enough to enable a real robot to manipulate the objects directly without adjusting any parameters.

Our robotic platform consists of a Franka Emika Panda arm, a wrist-mounted force/torque sensor, and a parallel gripper as the end-effector equipped with a Realsense RGB-D camera. For evaluation, we select two small revolute objects (key and window lock) for two reasons: First, their revolute axes are close to the end-effector while grasped. As a result even minor estimation errors can result in significant deviations in the articulation trajectory, causing large reaction forces. Second, we observes considerable tangent errors with these objects across all approaches (see TABLE~\ref{tab:baseline_comp_tab}).

\begin{figure}
	\centering
	\includegraphics[width=0.9\columnwidth]{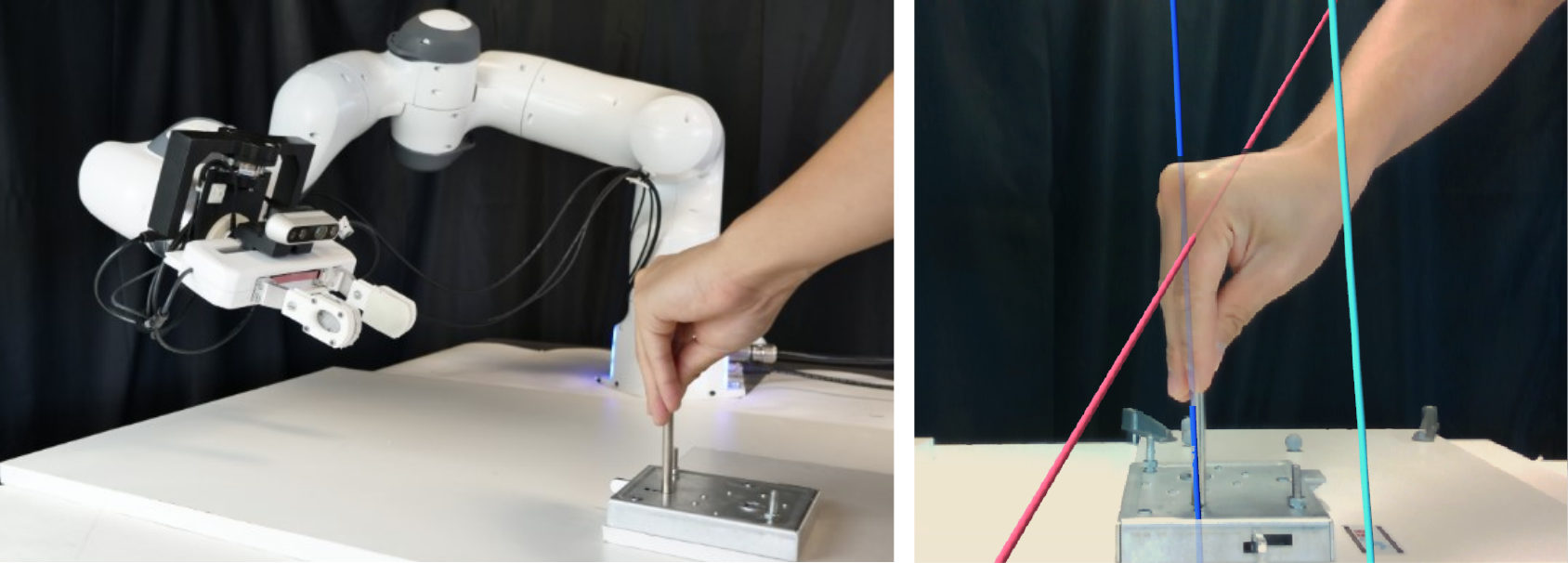}
	\caption{The left figure illustrates the robot estimating the kinematic model of the key by observing the motion of the human hand. The right figure shows that our approach (blue) achieves a more accurate estimation compared to baselines: Regal et al.~\cite{regal2023usingSingle} (pink) and Bahety et al.~\cite{bahety2024screwmimic} (cyan).}
	\label{fig:real_world_exp_setup}
    \vspace{-1em}
\end{figure}

In the experiment, the robot is placed in a predefined pose, observing how a human manipulates the object and estimating the kinematic model using the RGB-D camera, as shown in Fig.~\ref{fig:real_world_exp_setup}. After obtaining the kinematic model, the robot grasps the object, generates a Cartesian trajectory based on the estimated kinematic model, and tracks it using a Jacobian-transpose-based operational space controller~\cite{khatib1987unified}. We assume the grasp pose is known and use a visual servoing program to guide the end-effector to the predefined grasp pose~\cite{hutchinson1996tutorial}. We record five human manipulations, corresponding to five different estimates per method, and run a manipulation trial for each estimate. A trial is considered successful if the robot articulates the object over \SI{180}{\degree} while staying within a force magnitude threshold of \SI{70}{\newton}.

The success rates of all methods are summarized in TABLE~\ref{table:successrates}. Both baseline methods fail to manipulate the key in all trials due to the inaccuracy of the estimated models, which causes the robot to reach the force limit during manipulation. For the window lock, the approach by Regal et al.~\cite{regal2023usingSingle} completed the task three out of five times. The increased success rate is due to the non-rigid grasp between the end-effector and the grasped part of the window lock, allowing more relative rotations between both. These relative motions limit reaction forces during the manipulation. Consequently, Regal et al.~\cite{regal2023usingSingle} can complete the task, even with considerable errors in the rotation axis estimation. However, Bahety et al.~\cite{bahety2024screwmimic} struggle with generating accurate estimates of the rotation center, which generates significant reaction forces and leads to failure. 

\begin{figure}
	\centering
	\includegraphics[width=0.95\columnwidth]{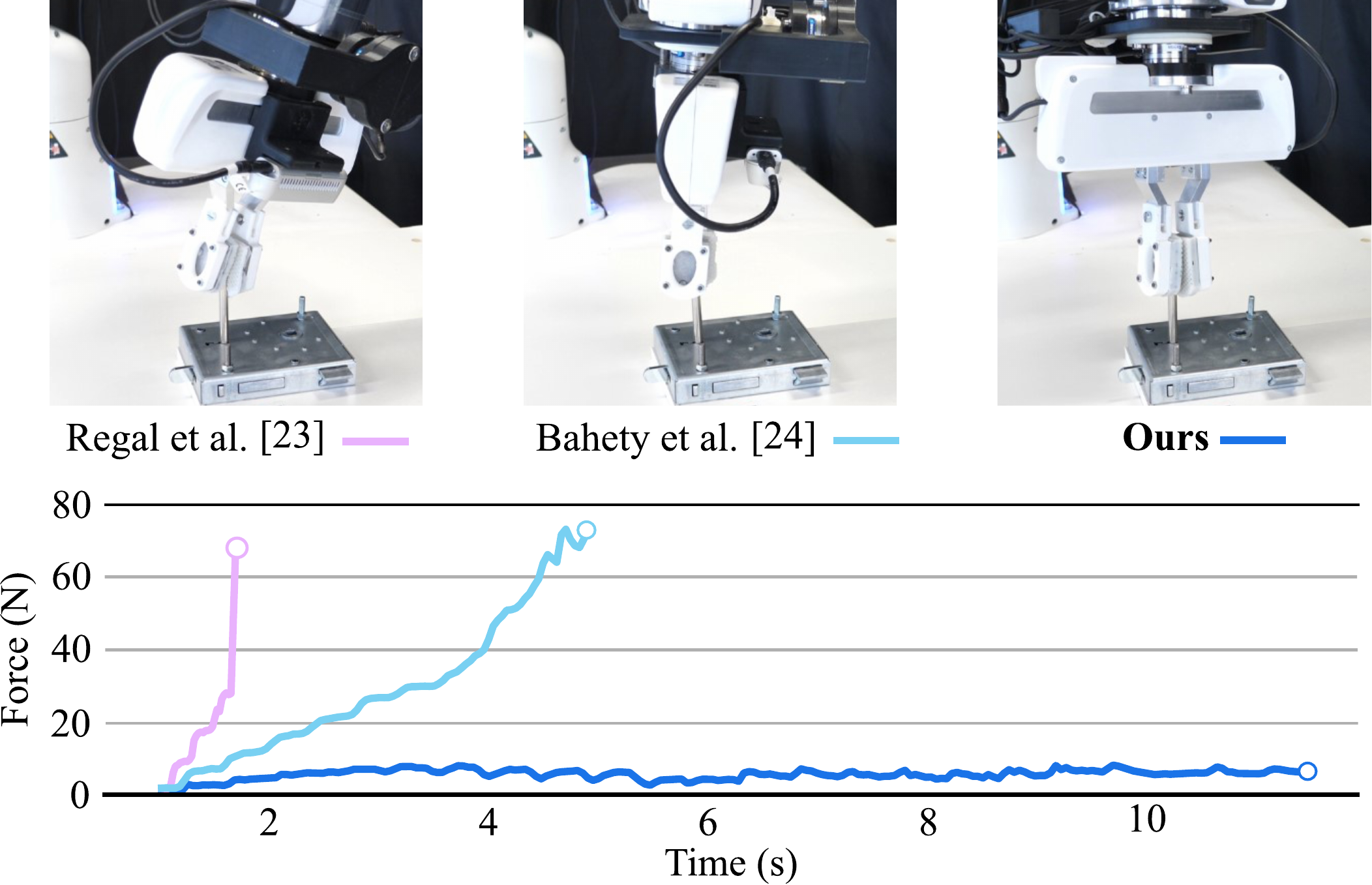}
	\caption{Force magnitude for the manipulation with the key using the estimated kinematic models shown in Fig.~\ref{fig:real_world_exp_setup}. The termination poses of the robot are shown above. While the two baselines reach the force limit during manipulation, our approach enables the robot to articulate the key over a wider motion range while observing significantly lower reaction forces.}
	\label{fig:forceplot}
	\vspace{-0.5em}
\end{figure}

Different to the baseline methods, our approach succeeds in all trials with both objects. Fig.~\ref{fig:forceplot} shows the force measurement's magnitude for all approaches in one manipulation trial with the key. While the baseline methods fail due to reaching the force limit or slippage, our approach successfully rotates both objects by \SI{180}{\degree} with an average force of \SI{12}{\newton} in all trials. Moreover, since our approach does not directly estimate hand mesh from images, it achieves averages at 22~FPS on a Nvidia A4000 GPU, showing real-time capability. These results strongly support that our approach efficiently and accurately estimates kinematic models from human hand motions, enabling a robot to directly manipulate the target object without the need for extensive interactive fine-tuning.

\begin{table}
	\centering
		\begin{tabular}{l c c c }
			\toprule
			Object  & Regal et al. \cite{regal2023usingSingle} & Bahety et al. \cite{bahety2024screwmimic} & \textbf{Ours} \\ 
			\midrule
			Key & 0/5 & 0/5 & \textbf{5/5}\\
			Window Lock &3/5 & 0/5 & \textbf{5/5} \\
			\bottomrule
		\end{tabular}
	\caption{Comparison of success rates for manipulating the key and window lock. Note that the method by Bahety et al.~\cite{bahety2024screwmimic} is evaluated without interactive fine-tuning on the robot.}
	\label{table:successrates}
	\centering
    \vspace{-0.8em}
\end{table}

\section{Limitations}
\label{sec:limitation}
Our approach has two main limitations. First, it assumes constant hand-object contact. Future work could address this by segmenting contact phases using likelihoods of the estimated kinematic models~\cite{niekum2015online} or contact detection methods~\cite{brahmbhatt2020contactpose, cho2024dense}. This would also enable our approach to estimate a sequence of kinematic models. Second, it requires an optimal viewpoint for full-sequence perception and reliable depth. To improve this, we plan to integrate a component that adjusts the robot's viewpoint for better depth sensing~\cite{ling2024articulated} and information gain~\cite{kiciroglu2020activemocap, breyer2022closed}. This could be achieved directly by optimizing the viewpoint via gradient descent through our estimation model~\cite{mengers2025noplan}.

\section{Conclusion}
\label{sec:conclusion}

We have developed a probabilistic method for real-time kinematic model estimation by tracking human hand motions. The experimental results demonstrate that our approach outperforms two offline baseline methods, providing highly accurate estimations for various articulated objects, including those with minimal articulation and substantial visual uncertainties. The precision of the estimated models is sufficient to ensure safe robotic manipulation. The strong performance of our approach underscores the value of using the human hand as a prior, as it enhances motion tracking and facilitates uncertainty modeling. 

\bibliographystyle{IEEEtran}
\bibliography{mybib}
\end{document}